\documentclass[10pt,journal,twoside,twocolumn]{article}

\usepackage[a4paper, left=0.5in, right=0.5in, top=0.75in, bottom=0.75in]{geometry}
\usepackage{times}
\usepackage[colorlinks=true,linkcolor=blue]{hyperref}%
\usepackage{amsmath,amssymb,amsfonts}
\usepackage{algorithmic}
\usepackage{graphicx}
\usepackage{xcolor}
\usepackage{subcaption}
\usepackage{multirow}
\usepackage{placeins}
\usepackage{booktabs} 
\usepackage{textcomp}
\usepackage[font=small,labelfont=bf]{caption}

\usepackage{titling}
\pretitle{\centering\vspace{1em}\rule{\textwidth}{0.5pt}\vspace{1em}\Huge\bfseries}
\posttitle{\par\vspace{1em}\rule{\textwidth}{0.5pt}\vspace{2em}}

\usepackage{sectsty}
\usepackage{xcolor}
\definecolor{darkblue}{rgb}{0,0.541,0.855}%
\sectionfont{\color{darkblue}}
\subsectionfont{\color{darkblue}}
\subsubsectionfont{\color{darkblue}}

\usepackage{fancyhdr}
\date{}

\renewenvironment{abstract}
  {\noindent\textbf{}\par\noindent\ignorespaces}
  {\par\noindent\ignorespacesafterend}

\begin{document}
\title{ \\Deep learning for temporal super-resolution \\4D Flow MRI}

\author{Pia Callmer,  Mia Bonini, Edward Ferdian, David Nordsletten, Daniel Giese, Alistair A. Young,  \\Alexander Fyrdahl, David Marlevi
\thanks{
\noindent
This work was funded in part by the European Union (ERC,
MultiPRESS, 101075494). Views and opinions expressed are those
of the authors and do not reflect those of the European Union or
the European Research Council Executive Agency. 
}
\thanks{
P.C., A.F., and D.M. are with Karolinska Institute, Solna, Sweden. 
M.B. and D.N. are with the University of Michigan, Ann Arbor, USA. 
E.F. is with Telkom University, Bandung, Indonesia and University of Auckland, Auckland, New Zealand.
D.G. is with Siemens Healthineers AG, Erlangen, Germany and Friedrich-Alexander-Universität Erlangen-Nürnberg, Erlangen, Germany.
A.A.Y. and D.N. are also with King’s College London, London, UK. 
A.F is also with Karolinska University Hospital, Solna, Sweden.
D.M. is also with Massachusetts Institute of Technology, Cambridge, USA}
}

\maketitle

\begin{abstract}
\textbf{4D Flow Magnetic Resonance Imaging (4D Flow MRI) is a non-invasive technique for volumetric, time-resolved blood flow quantification. However, apparent trade-offs between acquisition time, image noise, and resolution limit clinical applicability. In particular, in regions of highly transient flow, coarse temporal resolution can hinder accurate capture of physiologically relevant flow variations. To overcome these issues, post-processing techniques using deep learning have shown promising results to enhance resolution post-scan using so-called super-resolution networks. However, while super-resolution has been focusing on \emph{spatial} upsampling, \emph{temporal} super-resolution remains largely unexplored. The aim of this study was therefore to implement and evaluate a residual network for \emph{temporal} super-resolution 4D Flow MRI. To achieve this, an existing \emph{spatial} network (4DFlowNet) was re-designed for temporal upsampling, adapting input dimensions, and optimizing internal layer structures. Training and testing were performed using synthetic 4D Flow MRI data originating from patient-specific \emph{in-silico} models, as well as using \emph{in-vivo} datasets. Overall, excellent performance was achieved with input velocities effectively denoised and temporally upsampled, with a mean absolute error (MAE) of  1.0 cm/s in an unseen \emph{in-silico} setting, outperforming deterministic alternatives (linear interpolation MAE = 2.3 cm/s, sinc interpolation MAE = 2.6 cm/s). Further, the network synthesized high-resolution temporal information from unseen low-resolution \emph{in-vivo} data, with strong correlation observed at peak flow frames. As such, our results highlight the potential of utilizing data-driven neural networks for temporal super-resolution 4D Flow MRI, enabling high-frame-rate flow quantification without extending acquisition times beyond clinically acceptable limits.}
\end{abstract}

\section{Introduction}
Hemodynamic quantification is an important part of cardiovascular diagnostics, where regional variations in blood flow are directly coupled to cardiovascular performance \cite{Richter2006CardiologyFlow}. This holds particularly true for the heart itself, where cardiac output and long-term cardiac efficiency are directly linked to the complex and rapidly changing flow patterns observed in the heart \cite{Bolger2007TransitResonance}. Time-resolved three-dimensional phase-contrast magnetic resonance imaging, more commonly known as 4D Flow MRI, has emerged as a promising tool offering non-invasive quantification of full-field volumetric blood flow \cite{Markl20124DMRI} \cite{Zhuang2021ThePerspectives}. The technique has been explored in various cardiac settings, exemplifying how full-field cardiac flow mapping offers novel insights into both normal \cite{Ha20194DValves} and pathological cardiac behavior \cite{Vasanawala2015CongenitalMRI}. 
\\
4D Flow MRI refers to phase-contrast MRI with flow-encoding in three spatial directions, with reconstructed image data resolved over time, generating a four-dimensional data representation. The use of multiple encoding directions necessitates data acquisition over many cardiac cycles. The trade-off between acquisition time, and signal-to-noise-ratio (SNR) \cite{Markl20124DMRI} imposes a practical limit on attainable spatiotemporal resolution in a clinically feasible time frame. Whereas limitations in \emph{spatial} resolution have been highlighted when imaging narrow vascular structures \cite{Ferdian2023CerebrovascularMRI} \cite{Stankovic2014ReproducibilityResolution}, limitations in \emph{temporal} resolution have other implications when imaging the rapidly varying flow in the heart. Effects of coarse temporal resolution in 4D Flow MRI may manifest as reduced peak flow \cite{Carlsson2011QuantificationT} or net flow rates \cite{Hanneman2014MagneticImaging} compared to higher-temporal resolution 2D phase-contrast equivalents.
\\
In this context, several techniques have been proposed to either increase resolution or reduce acquisition time with maintained resolution. While novel protocols or sampling patterns promise enhanced acquisition efficiency \cite{Fyrdahl2020Sector-wiseDysfunction}, post-processing approaches attempt to improve the image quality of already acquired datasets. Several studies have explored the use of computational fluid dynamics (CFD) models, using geometries and boundary conditions extracted from image data \cite{Leuprecht2003BloodStudy} \cite{Bonini2022HemodynamicRegurgitation} to simulate patient-specific flow patterns. While enabling virtually unlimited spatiotemporal resolution, patient-specific CFD simulations directly depend on predefined boundary conditions and require high-performance computational resources not readily available in a clinical setting \cite{Dyverfeldt20154DStatement}. 
\\
Recent advances in machine learning have explored \emph{super}-resolved imaging, improving resolution, suppressing noise, and even filling in missing information in the input data \cite{Dong2014ImageNetworks}, and is now being explored for 4D Flow MRI enhancement. Specifically, two major approaches can be defined: using convolutional neural networks (CNN) to learn a direct mapping between low- and high-resolution image features or using function approximators to learn a continuous mapping between spatiotemporal coordinates and the equivalent velocity component. For the latter, both a purely data-driven neural implicit representation \cite{Saitta2024ImplicitMRI} and physical-informed neural networks (PINN) \cite{Fathi2020Super-resolutionNets} have been proposed as patient-specific functions. PINNs are characterized by an additional physics regularization term posed on the loss function to enforce a solution consistent with fluid-dynamic principles. In principle, the coordinate mapping nature of these approaches promises unlimited spatiotemporal resolution. However, the clinical applicability is limited due to requirement of boundary information and their patient-specific representation making pre-processing more cumbersome and computationally demanding re-training necessary for every new patient.
In contrast, traditional residual CNNs have been utilized to enhance 4D Flow MRI resolution across various domains \cite{Ferdian2023CerebrovascularMRI} \cite{Ferdian20204DFlowNet:Dynamics} \cite{Shit2022SRflow:Data} \cite{Long2023Super-resolutionLearning}, even exploring the use of ensemble CNN learning to generalize performance into \emph{unseen} flow domains \cite{Ericsson2024GeneralizedSystem}. However, these networks have been exclusively utilized to enhance \emph{spatial} resolution, leaving temporal upsampling an unsolved problem.
\\
The aim of this study is therefore to develop a first-of-a-kind \emph{temporal} super-resolution network for 4D Flow MRI applicable for diverse patient geometries without the need for patient-specific re-training, enabling accurate capture of rapid flow phenomena with a clinically feasible scan time. Specifically, by re-designing an existing \emph{spatial} super-resolution CNN for \emph{temporal} upsampling, we leverage previous advances to enable frame rate representations beyond current practice, with capabilities allowing for temporal flow recovery across various cardiac anatomies. Focusing on left ventricular flow dynamics, training, validation, and testing are performed with both synthetically generated 4D Flow MRI data as well as \emph{in-vivo} data to ensure broad applicability of our proposed model.  
Overall, our contributions lie in \emph{(1)} re-designing and optimizing a core CNN setup for \emph{temporal} super-resolution imaging, 
\emph{(2)} developing a data processing pipeline to mimic temporal sampling patterns in synthetic training data, and \emph{(3)} assessing performance on experimentally acquired \emph{in-vivo} data, showcasing translation into a direct clinical setting.
\label{sec:introduction}
\section{Methods}
\subsection{Network model}
\subsubsection{Temporal super-resolution architecture}
\begin{figure*}[!ht]
    \centering
    \includegraphics[width=\textwidth, alt={Illustration of the proposed temporal super-resolution network for 4D Flow MRI data.}]{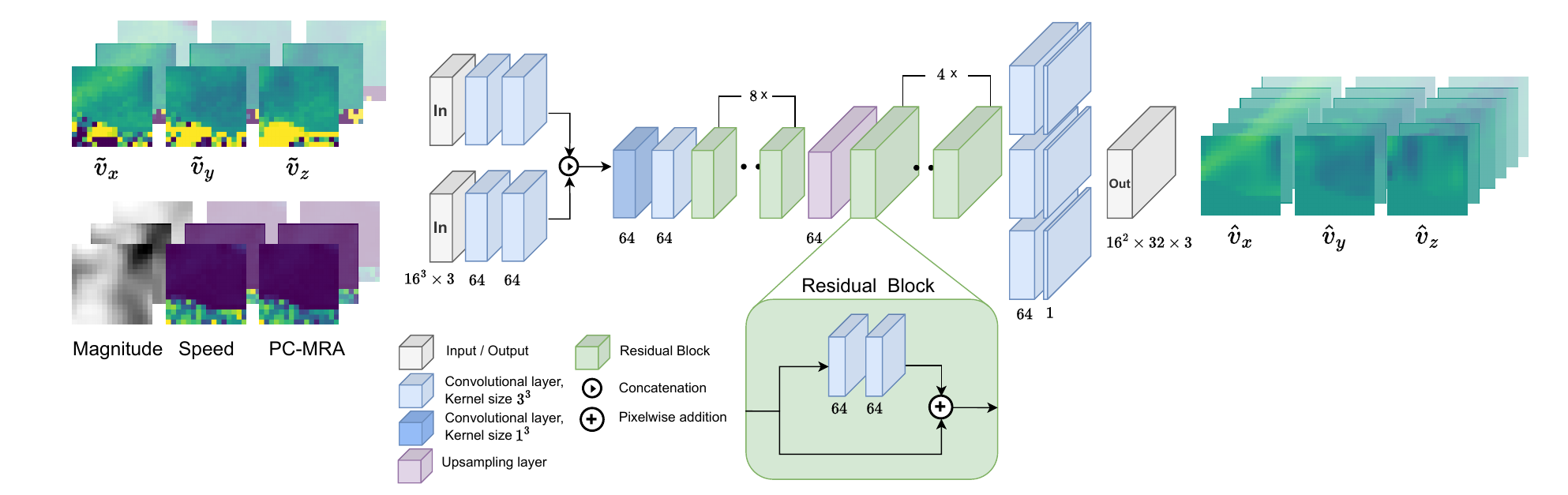}
    \caption{Illustration of the proposed temporal super-resolution network for 4D Flow MRI, based on the previously described 4DFlowNet \cite{Ferdian20204DFlowNet:Dynamics}. The residual convolutional neural network takes a sequence of 2D low resolution, noisy input velocities $\Tilde{v}_x$, $\Tilde{v}_x$, $\Tilde{v}_x$, magnitude, speed and a phase-contrast magnetic resonance angiogram (PC-MRA) mask, and outputs denoised and super-resolved velocities $\hat{v}_x$, $\hat{v}_x$, $\hat{v}_x$.   
    }
    \label{fig:Methods_arch}
\end{figure*}
As a starting point for creating a \emph{temporal} super-resolution network, the convolutional \emph{spatial} super-resolution network 4DFlowNet \cite{Ferdian2023CerebrovascularMRI}\cite{Ferdian20204DFlowNet:Dynamics} was used as the base architecture. Primary architectural adaptations of 4DFlowNet for temporal upsampling were required in the input and output layers, along with alterations to the central upsampling layer. The original spatial network acts on 3D \emph{spatial} patches at static time points, with 3D convolutions used to pass information through the network. To maintain this network structure for temporal upsampling, we opted for two-dimensional spatial slices sampled over time (2D+t), creating 3D patches conforming with the original network structure. These alterations also necessitated changing the central upsampling layer from tri-linear to linear upsampling, enhancing resolution in the temporal dimension only. Beyond the core architecture, activation functions, kernel sizes, and network depth were kept identical. An overview of the modified network is given in Fig. \ref{fig:Methods_arch}.

\subsubsection{Loss function}
The network was trained to minimize velocity magnitude and velocity direction differences between predicted super-resolved and reference high-resolution data. This was achieved using a combined loss function, minimizing a mean squared error (MSE) loss together with a previously proposed mutually projected loss \cite{Shit2022SRflow:Data}. 
\\
The MSE between high resolution velocity $\mathbf{v}_i = [v_x, v_y, v_z] \in \mathbb{R}^3$ and super-resolved velocity $\hat{\mathbf{v}}_i \in \mathbb{R}^3$ with $i \in \{1, .., n\}$, was defined as 
\begin{equation*}
    L_{\text{MSE}}(\mathbf{v}, \hat{\mathbf{v}}) = \frac{1}{n}  \sum_{i=1}^n || \mathbf{v}_i - \hat{\mathbf{v}}_i||^2_2,
\end{equation*}
where $n$ is the total number of voxels in a given patch. For the MSE loss, this was further separated into fluid and non-fluid regions as
\begin{equation*}
    L_\text{MSE-total}(\mathbf{v}, \hat{\mathbf{v}}) = L_\text{MSE-fluid} (\mathbf{v}, \hat{\mathbf{v}}) + L_\text{MSE-nonfluid} (\mathbf{v}, \hat{\mathbf{v}}).
\end{equation*}

The mutually projected loss was then defined within the fluid region as the convex linear combination of two projected $L_1$ losses, given by 
\begin{align*}
    L_\text{mp-l$_1$}(\mathbf{v}, \hat{\mathbf{v}}) &= \frac{1}{n_f} \sum_{i=1}^{n_f} \left( \beta L_{1_\text{I}}(\mathbf{v}_i, \hat{\mathbf{v}}_i) + (1-\beta) L_{1_{\text{II}}}(\mathbf{v}_i, \hat{\mathbf{v}}_i) \right), \\ \quad \text{where } & \beta \in [0, 1]
\end{align*}
with 
\begin{equation*}
    L_{1_\text{I}}(\mathbf{v}_i, \hat{\mathbf{v}}_i) = |||\mathbf{v}_i|| - || \hat{\mathbf{v}}_i|| \cos(\theta)|
\end{equation*}
and 
\begin{equation*}
    L_{1_{\text{II}}}(\mathbf{v}_i, \hat{\mathbf{v}}_i) = ||| \hat{\mathbf{v}}_i|| - ||\mathbf{v}_i || \cos(\theta)|
\end{equation*}
where $\theta$ is the angle between $\mathbf{v}_i$ and $\hat{\mathbf{v}}_i$ and $n_f$ the number of voxels within the fluid region. Note that the purpose of adding a directional loss term was to increase the directional consistency in predicted velocity components.
Combining the two loss functions, the final loss function was defined as
\begin{align*}
    L_\text{total} (\mathbf{v}, \hat{\mathbf{v}}, \mathbf{w}) &= \alpha L_\text{MSE-total}(\mathbf{v}, \hat{\mathbf{v}}) + (1 - \alpha) L_\text{mp-l$_1$}(\mathbf{v} , \hat{\mathbf{v}})  \\ & + \lambda_\text{nn} ||\mathbf{w}||_2^2, 
\end{align*}
where $\alpha \in [0,1]$ determines the relative weight of the MSE and directional loss terms. Note also that an L$_2$ regularization term weighted by $\lambda_\text{nn}$ is introduced to penalize large network weights $\mathbf{w}$  to prevent overfitting (in our instance, $\lambda_\text{nn} = 5 \cdot 10^{-7}$). 

\subsubsection{Model training and implementation details}
The network was implemented in TensorFlow 2.9  and was trained for 160 epochs using the Adam optimizer \cite{Kingma2015Adam:Optimization} with an initial learning rate of $2 \cdot 10^{-4}$ and a batch size of 32. The parameters of the loss function were $\alpha = 0.8$ and $\beta = 0.5$. The training time was 44-48 hours on a single NVIDIA A100 Tensor Core GPU (Santa Clara, CA, USA). The model and code to reproduce the training has been made publicly available at \url{https://github.com/PiaaCaa/Temporal4DFlowNet}. 

\subsection{Training and testing data}
\subsubsection{Patient specific computational fluid dynamics models} \label{sec:meth_pat-specific}
To train a supervised temporal super-resolution network, paired data of temporal low and high-resolution is required. While perfectly matched pairs of 4D Flow MRI acquired at two different temporal resolutions would represent an ideal training set, it is in practice difficult to obtain high-resolution, high-SNR, ground-truth data, especially coupled to a perfectly matched low-resolution equivalent. To overcome this, we  opted for paired datasets of \emph{synthetic} 4D Flow MRI, derived from CFD models of real patient anatomies, replicating physiologically feasible flow patterns, as previously described \cite{Bonini2022HemodynamicRegurgitation}. The models were based on computed tomography (CT) images covering the left atrium, left ventricle, and aortic outflow. Six datasets derived from different patients exhibiting varying degrees of mitral regurgitation were used. With nodal output generated from the simulations, data was projected onto a voxelized image-equivalent grid of 2 mm isotropic, using finite element base functions for projection. This spatial projection was performed at a temporal discretization of 10 ms. 

\begin{figure}[!ht]
    \centering
    \includegraphics[width=0.5\textwidth, alt={Overview of in-silico data and patch generation for one of six data models. Shows voxelized geometry, velocity field, and patch selection.}]{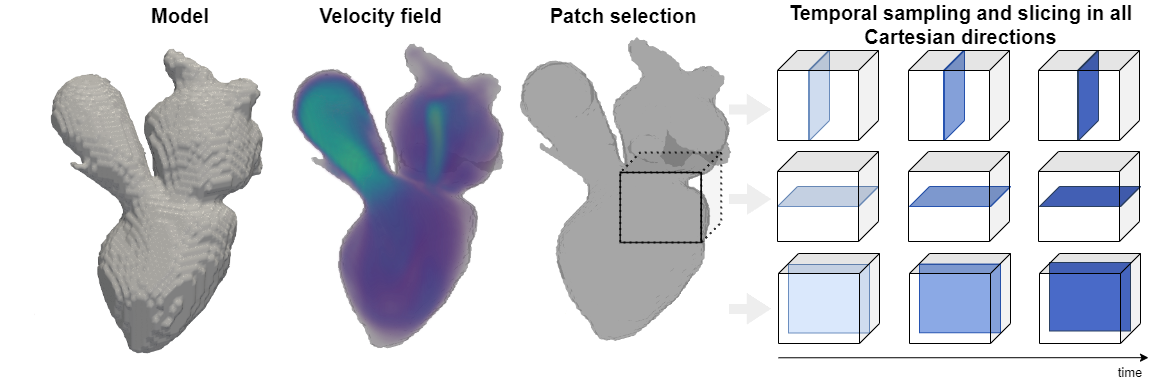}
    \caption{Overview of \emph{in-silico} data and patch generation for one of six data models. Left to right: voxelized geometry (2 mm resolution), velocity field (0–68 cm/s), and patch selection. Patches generate 2D slice sequences ($2D +t$) in all three Cartesian directions across multiple time frames.}
    \label{fig:methods_patchselection}
\end{figure}

\begin{figure*}[!ht]
    \centering
    \includegraphics[width=\textwidth, alt={Overview of the synthetic MRI data preparation pipeline. Steps include combining phase and magnitude images, generating coil sensitivity maps, adding noise in k-space, undersampling k-space data, compressed sensing reconstruction, and extracting velocity and magnitude data.}]{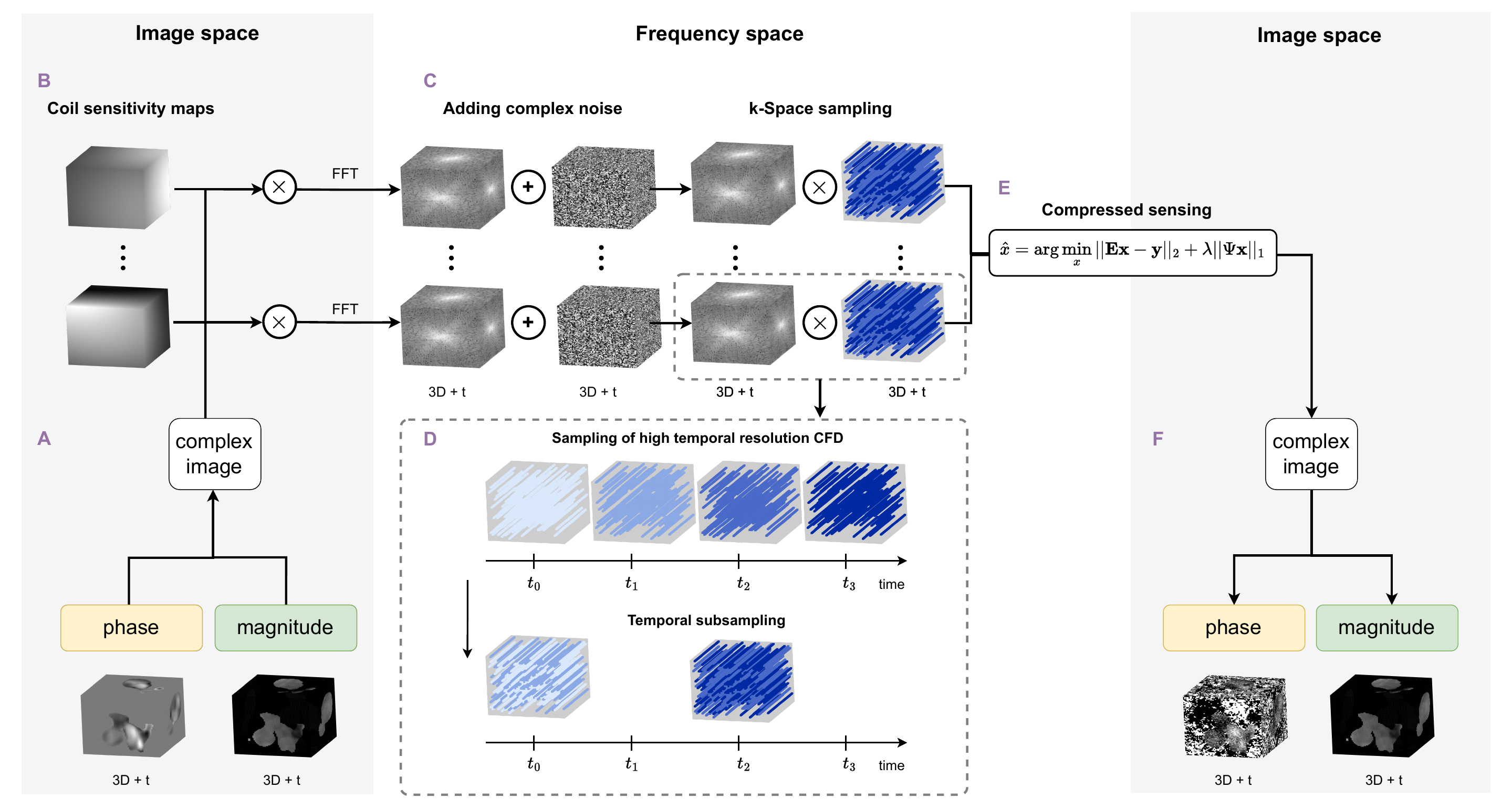}
    \caption{Overview of the synthetic MRI data preparation pipeline. (\textbf{A}) Combination of phase and magnitude images into a complex image. (\textbf{B})  Coil sensitivity maps are generated from Biot-Savart simulations, each of which is multiplied by the complex image. (\textbf{C}) Application of FFT and addition of complex Gaussian noise in the k-space. (\textbf{D}) Sampling of frequency domain data with a Cartesian variable-density phyllotaxis sampling pattern. Temporal subsampling is achieved by accumulating the readout pattern for consecutive frames. (\textbf{E}) Compressed sensing reconstruction of the undersampled k-space data (\textbf{F})  Extraction of velocity and magnitude data from the reconstructed image.}
    \label{fig:Methods_CFDMRI}
\end{figure*}

\subsubsection{Generating synthetic 4D Flow MRI data}
\label{sec:meth_insilico_generation}
With simulated velocity data obtained as per above, we implemented a data-processing pipeline to convert synthetic CFD data into realistic \emph{synthetic} 4D Flow MRI datasets to use as low resolution input data. 
This was performed to closely mimic the characteristics and temporal data processing of a real MR scanner to ensure that training was performed on realistic input data (see Fig. \ref{fig:Methods_CFDMRI}). Below follows a review of the steps taken:

\begin{enumerate}
    \item The uniformly sampled simulated CFD velocity $v_v$ is transformed into a corresponding phase velocity given by $v_p = (v_v/ \text{VENC}) \cdot \pi $ with $\text{VENC}$ an assigned velocity encoding. Together with a coupled magnitude image $m$, a complex value signal is given by $v_c = m \cdot e^{i\cdot v_p}$.
    \item  To simulate the spatial varying receiver field of an MRI scanner, coil sensitivity maps were simulated using a Biot-Savart model. The locations of the receiver coils were equally distributed around the model geometry. In total, 64 coil sensitivity maps were calculated with 8 randomly assigned as active for a given model. The sensitivity maps are then multiplied with the complex image space data.
    \item Data was converted into k-space using a Fast Fourier Transformation, after which complex noise is added at a target signal-to-noise ratio (SNR) randomly assigned between 14 - 17 dB or 40 - 45 db (two separate datasets).  
    \item To resemble the sampling pattern of an actual phase-contrast acquisition, a pseudo-random Cartesian variable-density phyllotaxis sampling pattern taken from a reference 4D Flow MRI sequence \cite{Forman2014High-resolutionCompensation} was used to sample data in k-space. 
    \item Data is then transformed back into image space using a compressed sensing reconstruction \cite{Lustig2007SparseImaging}.
    Specifically, this meant solving 
    \begin{equation*}
        \hat{x} = \arg \min_{x} ||\mathbf{E}\mathbf{x} - \mathbf{y} ||_2 + \lambda_\text{cs}||\Psi \mathbf{x}||_1,
    \end{equation*}
    where $\mathbf{x}$ is the image, $\mathbf{E}$ the encoding matrix encompassing coil sensitivities, the sampling mask, and the Fourier transform, $\mathbf{y}$ the sampled \emph{k}-space data, $\lambda_\text{cs}$ a regularization parameter and $\Psi$ the Haar wavelet transform. The reconstruction was optimized using the fast iterative shrinkage--thresholding algorithm (FISTA) \cite{Beck2009AProblems} implemented in the Berkeley Advanced Reconstruction Toolbox (BART) \cite{Uecker2015BerkeleyToolbox}. 
    \item The reconstructed image is then separated into velocity and magnitude representations using the same velocity encoding parameter as in step 1. 
\end{enumerate}

Note that for the above, the CFD-simulated velocity data was paired with magnitude images obtained from reference \emph{in-vivo} acquisitions, with these datasets rigidly co-registered to align the left ventricle in simulated and acquired datasets, similar to previous work \cite{Ferdian2023CerebrovascularMRI}.   

\subsubsection{Patch generation and data augmentation}
To generate a larger number of training sets from the six input models, as well as to minimize domain geometry dependence, the input data domain was split into several smaller patches. Spatial patches of $16 \times 16$ voxels were extracted from each model geometry, with each slice sampled over time to create low-resolution $2D+t$ input patches of $16 \times 16 \times 16$ voxels. High-resolution data was then patched equivalently, extracting data from the same spatial position, however, sampled at double the frame rate, generating patches of $ 16 \times 16 \times 32$ voxels. To increase the amount of training data, patches were extracted with 2D spatial slicing performed in all three Cartesian directions (see Fig. \ref{fig:methods_patchselection}). The patches were selected from random positions within the model geometry such that each patch contained at least $20\%$ voxels with fluid content over time. Additionally, one patch with less than $20\%$ fluid content was included in each sampling iteration, seeking to make the network more robust to non-fluid regions.
Further, to avoid directional bias, we generated additional training samples for each patch by randomly applying a spatial rotation of either 90, 180, or 270 degrees, flipping patches horizontally or vertically, changing velocity to opposite direction or swapping velocity components, as well as including either low-noise patches (target SNR of 40-45 db) or higher noise patches (target SNR of 14-17 dB). The validation and test set included high-noise data (target SNR of 14-17 dB) and no data augmentation was performed.
\\
Following the above, we ended up with 66'692 patches for training, 4'346 patches for validation, and 4'244 patches for testing. The division was performed on the model level, ensuring that any single input model was only present in either the training, validation, or testing set, respectively. Once trained, the inference was performed on a patch-based level, with input data split into patches of the aforementioned sizes. The patches were automatically stitched together, using an overlap of 4 voxels, and temporal estimates were averaged across 2D slices from all Cartesian directions. 

\subsubsection{\emph{In-vivo} data acquisition}
To complement assessments performed on the synthetic datasets, actual 4D Flow MRI data was also utilized for evaluation. Specifically, two sets of data with different acquisition parameters, as detailed in Table \ref{tab:meth_mriparams}.

\underline{\textbf{Recovery to native resolution}}:
\label{sec:methods_invivo_nativeres}
First, 4D Flow MRI was acquired at a fixed temporal resolution in five (n=5) healthy volunteers 
at 1.5 T (MAGNETOM Sola, Siemens Healthineers, Forchheim, Germany). 4D Flow MRI was acquired using a research pulse sequence with a field-of-view covering the entire heart, a spatial resolution of 3 mm$^3$ isotropic, and a fixed temporal resolution of 33.4 ms. As a first attempt to assess recovery of unknown temporal frames, we created a low-resolution counterpart by synthetically downsampling the acquired data in time by removing every second temporal frame, reducing the temporal resolution to 66.8 ms. 

\underline{\textbf{Paired data acquisition}}:
\label{sec:methods_invivo_paireddata}
Second, to increase complexity and get closer to clinical temporal super-resolution, paired low and high-temporal resolution 4D Flow MRI was also acquired in five (n=5) healthy volunteers (mean age: $28.6 \pm 4.1$ years, two females, three males) at 1.5 T (MAGNETOM Sola, Siemens Healthineers, Forchheim, Germany) using a research pulse sequence. 4D Flow MRI was again acquired with a field-of-view covering the entire heart, this time using a spatial resolution of 2.5 mm$^3$ isotropic and a temporal resolution of both 20 and 40 ms. Note that the two 4D Flow MRI acquisitions were performed sequentially in the same scan session.  The acquisition times were approximately 15--20 minutes for the low temporal resolution datasets, and 30--40 minutes for the high temporal resolution datasets, depending on the heart rate and breathing pattern of the volunteers. 
All \emph{in-vivo} data were acquired with informed consent and approval from the Swedish ethical review board.

\begin{table}[h]
    \centering
    \caption{MRI sequence parameters. Dataset II includes both low- and high-temporal resolution datasets, with different parameters for both temporal resolution and the number of segments.  $^1$ FOV: Field-of-view; TR: repetition time; TE: echo time; ECG: electrocardiogram.}
    \renewcommand{\arraystretch}{1.2}
    \resizebox{0.5\textwidth}{!}{
    \begin{tabular}{l c c }
    \hline
    &                               \textbf{Dataset I} & \textbf{Dataset II} \\
     \hline  
         Field strength [T]         & 1.5 & 1.5 \\
         Spatial resolution [mm$^3$]    & 3                   & 2.5             \\
         Temporal resolution [ms]   &  33.4                 &     20, 40              \\
         Velocity encoding [cm/s]   & 150--175                 &  175--200        \\
         FOV [mm$^3$]                   &   480 $\times$ 324 $\times$ 144    &    400 $\times$ 325 $\times$ 180                 \\
         Matrix size [px]           &           160 $\times$ 108 $\times$ 40      &   160 $\times$ 128 $\times$ 72                   \\
         TE / TR [ms]               & 2.24 / 4.18&2.81 / 5.02  \\
         Flip angle [°]             & 8 & 8  \\
         Gating                     & Retrospective  &  Retrospective \\
         Acceleration type          & Compressed sensing    & Compressed sensing  \\
         Acceleration factor        & 7.7               & 7.6    \\
         Readout bandwidth [Hz/px]  &    558                   & 460 \\
         Segmentation factor        & 2                     & 1, 2 \\
         \hline \hline
    \end{tabular}}
    \label{tab:meth_mriparams}
\end{table}

\subsection{Performance evaluation}
 \subsubsection{In-silico validation}
 \label{sec:methods_insilico_validation}
 To evaluate the performance of the temporal super-resolution network, synthetic 4D Flow MRI data from Section \ref{sec:meth_insilico_generation} were used. Consistently, performance was evaluated by comparing super-resolved velocities to high-resolution references both during peak early diastole, as well as across the entire simulated cardiac cycle. The evaluations were performed on a separate test model, distinct from the models used during training and validation. 

 \subsubsection{In-vivo verification}
 \label{sec:methods_invivo_verification}
 \underline{\textbf{Recovery to native resolution}}:
In addition to the validation described above, the synthetically downsampled \emph{in-vivo} datasets from Section \ref{sec:methods_invivo_nativeres} were returned to their native temporal resolution by the super-resolution network. The validation was performed within the fluid domain defined by a manually delineated segmentation mask covering the left ventricle and aortic arch, respectively.

 \underline{\textbf{Paired low- and high-resolution data}}:
 Network evaluation was also performed on the paired low- and high-resolution datasets, comparing high-resolution reference velocities to super-resolved equivalents derived from the low-resolution input data. As with the previous \emph{in-vivo} data, manually delineated segmentation of the left ventricle and aorta were used to define the validation region, where the very same segmentation mask was used for both low- and high-resolution data.

\begin{figure*}[!ht]
    \centering
    \includegraphics[width=\textwidth, alt={In-silico evaluation on the test set. Shows comparison of $V_y$ between low resolution, high-resolution, super-resolution, and sinc interpolation during selected time frames at early diastole; linear regression plot at the peak synthesized frame of early diastole; and evolution of linear regression parameters $k$ and $R^2$ over the cardiac cycle.}]{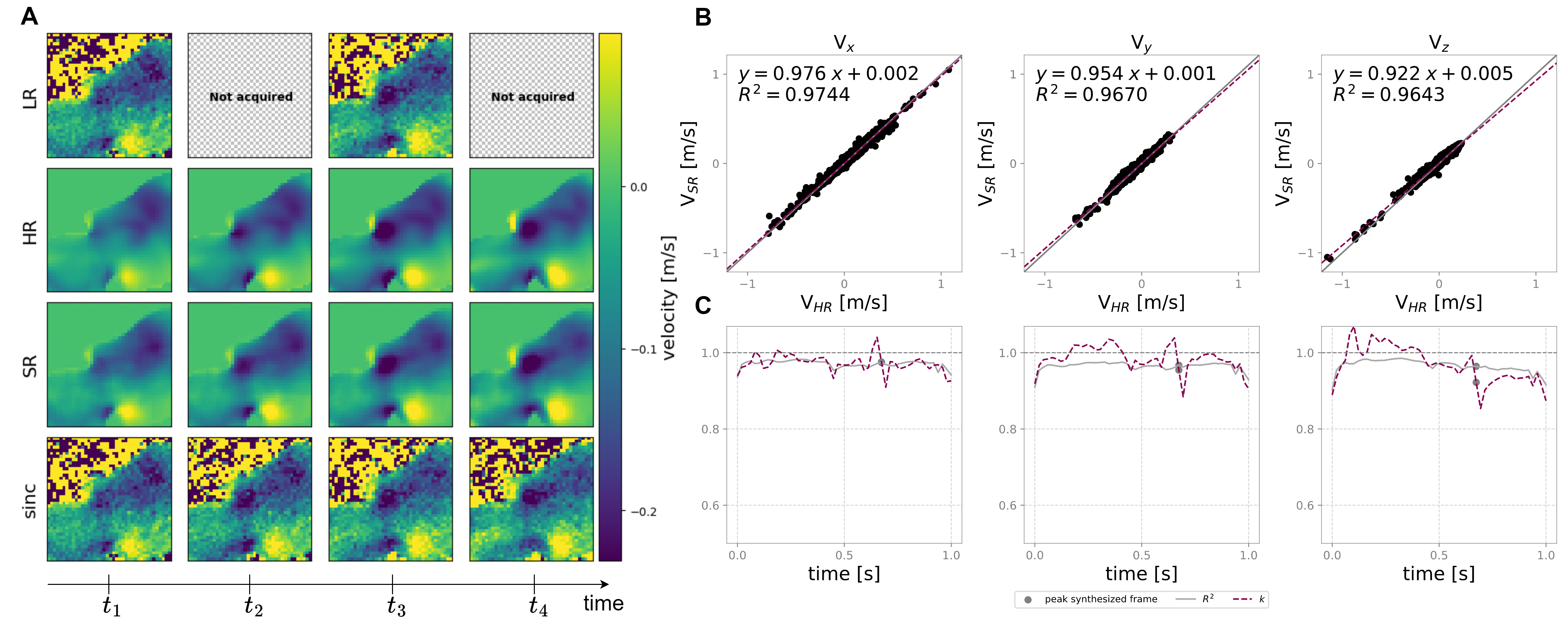}
    \caption{ \emph{In-silico} evaluation on the test set. From left to right: \textbf{(A})
    Comparison of $V_y$ between low resolution (LR), high-resolution (HR), super-resolution (SR) and sinc interpolation during a few selected time frames at early diastole. \textbf{(B}) Linear regression plot for one of the peak synthesized frames at early diastole. (\textbf{C}) Evolution of linear regression parameters $k$ and $R^2$ over the cardiac cycle. 
    }
    \label{fig:results_insilico_I}
\end{figure*}

\begin{table*}[btp]
    \centering
    \caption{Evaluation of super resolution, sinc interpolation, and linear interpolation methods on the fluid and boundary regions of the \emph{in-silico} test set. The metrics are averaged over all time frames and represent the average over the whole region. The best performance for each metric and in each region is marked with boldface.
    Each metric has the following units: RMSE [m/s], MAE [m/s], MRE [\%].}
    \resizebox{\textwidth}{!}{\begin{tabular}{cccccccc}
    \hline
Region & Method & RMSE & $|1-k|$& $R^2$ & MAE  & MRE & Cosine sim. \\ \hline \hline
\multirow{3}{*}{Fluid} 
&SR       & (\textbf{0.015},  \textbf{0.013}, \textbf{0.014}) & (\textbf{0.030}, \textbf{0.027}, \textbf{0.047}) &   (\textbf{0.971}, \textbf{0.964}, \textbf{0.965}) & (\textbf{0.011},  \textbf{0.009}, \textbf{0.010}) & \textbf{24.2} & \textbf{0.95} \\
& linear  & (0.033, 0.028, 0.030) & (0.082, 0.060, 0.088) &  (0.869, 0.834, 0.841) & (0.025, 0.022, 0.022) & 45.0 & 0.84 \\
& sinc    & (0.040, 0.037, 0.037) & (0.091, 0.068, 0.096) &  (0.816, 0.745, 0.764) & (0.028, 0.025, 0.025) & 47.7 & 0.82 \\ \hline
\multirow{3}{*}{Boundary} &
SR       &  (\textbf{0.019}, \textbf{0.015}, \textbf{0.016}) &  (\textbf{0.085} , 0.116, 0.067) &   (\textbf{0.893}, \textbf{0.873},  \textbf{0.919}) & (\textbf{0.013}, \textbf{0.011}, \textbf{0.011}) & \textbf{45.8} & \textbf{0.83}\\
&linear  &  (0.043, 0.037, 0.039) &  (0.095, \textbf{0.076}, \textbf{0.057}) &  (0.604, 0.514, 0.655) & (0.032, 0.027, 0.029) & 69.1 & 0.63\\
&sinc    &  (0.062, 0.062, 0.062) &  (0.110, 0.086, 0.073) &  (0.454, 0.345, 0.485) & (0.041, 0.039, 0.040) & 72.8 & 0.58\\ \hline \hline
\end{tabular}} 
    \label{tab:results_insilico_SR_linear_sinc_fluid_b_nonfluid}
\end{table*}

\subsubsection{Error metrics}
In all of the above, network performance was assessed by comparing high-resolution velocities $\mathbf{v}$ to its super-resolution counterparts $\hat{\mathbf{v}}$. Relative error, mean squared error (MSE), and cosine similarity were assessed in voxels within a pre-defined fluid region or segmentation. Note that the relative error was defined as 
\begin{equation}
    \text{RE} (\mathbf{v}, \hat{\mathbf{v}}) = \frac{1}{n_f} \sum_{i=1}^{n_{f}} \tanh \left( \frac{||\mathbf{v}_i - \hat{\mathbf{v}}_i||_2}{||\mathbf{v}_i||_2} \right)
\end{equation}
where $n_f$ is the number of voxels within the fluid region, and where the tangens hyperbolicus is used to ensure a balanced weighting between high and low velocities. 
 Further, linear regression analysis was performed between super- and high-resolution data, reporting the regression slope ($k$) and coefficient of determination ($R^2$) in both peak flow frames and as a function of time. Note that we chose the absolute deviation from 1, i.e. $|1-k|$ as an unbiased error metric when averaging over time.
 Lastly, network performance was benchmarked against deterministic linear or sinc interpolation. 

\section{Results}
As noted in Section \ref{sec:methods_insilico_validation}, the network performance was initially evaluated on an unseen \emph{in-silico} model, converting low-resolution (LR) input data at a temporal resolution of 40 ms to super-resolution (SR) equivalents at a temporal resolution of 20 ms.

\begin{figure*}[ht!]
    \centering
\includegraphics[width=\textwidth, alt = {Flow evaluation over aortic and mitral valve planes of the in-silico test set. Shows qualitative comparison of LR, HR, and SR results for aortic and mitral valve planes; and mean velocity across outlets plotted as a function of time.}]{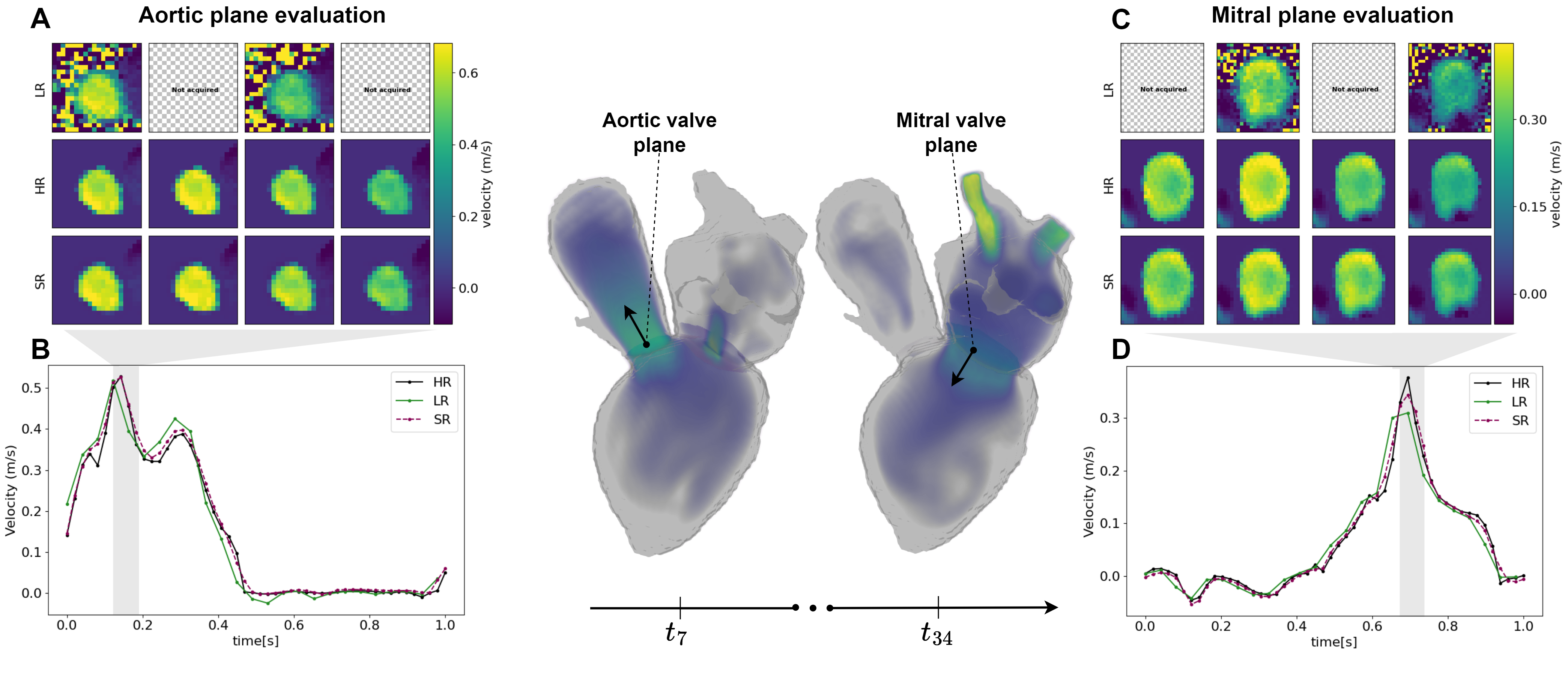}
    \caption{Flow evaluation over the defined aortic and mitral valve planes of the \emph{in-silico} test set. Qualitative evaluation comparing LR, HR and SR results for aortic and mitral valve planes (\textbf{A}, \textbf{C}). Mean velocity [m/s] across outlets plotted as a function of time (\textbf{B}, \textbf{D}). } 
    \label{fig:results_insilico_AV_MV}
\end{figure*}

\begin{figure*}[btp]
    \centering
    \includegraphics[width=\textwidth, alt ={Qualitative comparison of $V_x$ for subject A1 between LR, HR, and SR in peak systolic frames. Includes correlation plots between SR and HR for each synthetic frame prediction, covering all voxels within the left ventricle and aorta.}]{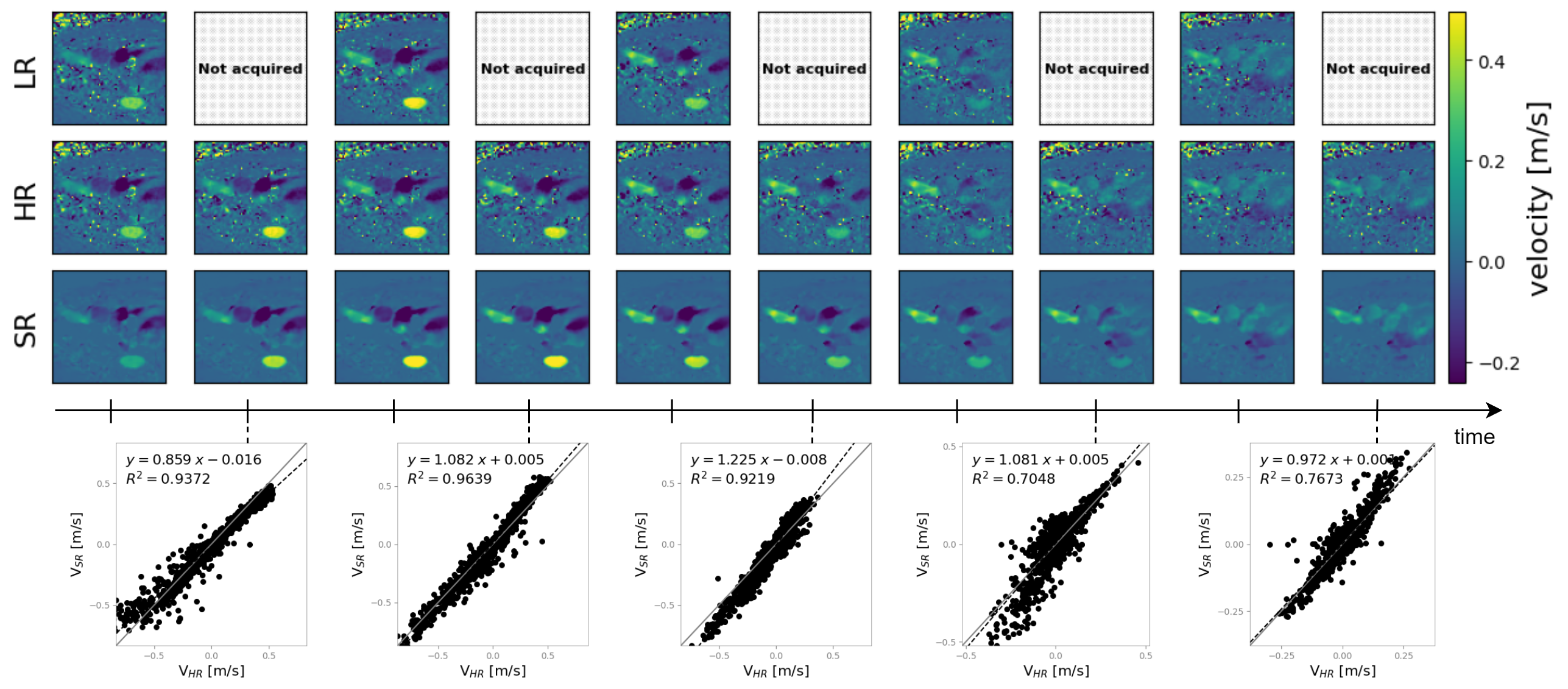}
    \caption{Qualitative comparison of $V_x$ for subject A1 between LR, HR and SR in peak systolic frames and the correlation plots between SR and HR for each synthetic frame prediction. The correlation plots include all voxels within the left ventricle and aorta. 
    }
    \label{fig:results_invivoI_qual_P01}
\end{figure*}

\subsection{In-silico validation}
As shown in Fig. \ref{fig:results_insilico_I}A and supplementary Video I, the network exhibits a substantial noise reduction compared to the low-resolution input, both inside and outside the fluid region. Further, the results indicate a successful synthesis of information contained in the reference high-resolution data absent in the low-resolution input. 
\\
The network achieves highly accurate predictions across the entire domain and all velocity directions with an average RMSE of 1.4 cm/s inside the fluid region, and 0.2 cm/s outside the fluid region. A close to perfect correlation to reference high-resolution data is observed, with an average linear regression deviation below 7\% (Table \ref{tab:results_insilico_SR_linear_sinc_fluid_b_nonfluid}). This performance is also visually depicted in Fig. \ref{fig:results_insilico_I}B-C, where linear regression between HR and SR in a synthesized frame during early diastole is shown for all velocity components, along with regression slope and coefficient of determination across the entire cardiac cycle.
\\
Isolating synthesized peak flow frames, slightly higher errors were observed with the network achieving an RMSE of 1.7 cm/s in early-diastole and 1.5 cm/s in peak systole, compared to 1.4 cm/s across the entire cardiac cycle. These errors were notably smaller than those from deterministic approaches: linear interpolation resulted in RMSEs of 3.4 cm/s (early-diastole), 3.2 cm/s (peak systole), and 3.0 cm/s (whole cycle), while sinc interpolation yielded higher RMSEs of 4.4 cm/s, 4.9 cm/s, and 3.8 cm/s, respectively.

\noindent
Complementing the above, Fig. \ref{fig:results_insilico_AV_MV} shows flow across the mitral and aortic valve planes, depicting velocities in peak systole across the aortic valve, and early diastole for the mitral valve. While local variations are apparent,  the network demonstrates an ability to synthesize data not present in the low-resolution input across both valve planes, with an error at peak flow below 0.21 cm/s for aortic and below 3.3 cm/s for mitral valve plane. 

\begin{table*}[ht]
\caption{Evaluation of network performance for reconstruction to native resolution on \emph{in-vivo} dataset I averaged over time (left) and at peak synthesized frame (right). }
    \centering
    \resizebox{\textwidth}{!}{
    \begin{tabular}{ccccccc}
            \hline 
            & \multicolumn{3}{c}{Mean performance} & \multicolumn{3}{c}{Peak synthesized frame} \\
             &  RMSE & $|1-k|$ & $R^2$ & RMSE & $k$ & $R^2$ \\
             \cmidrule(lr){2-4} \cmidrule(lr){5-7}
            A1 & (0.05, 0.04, 0.05) &  (0.21 , 0.32, 0.18) & (0.77, 0.68, 0.80) & (0.07, 0.05, 0.08) &  (0.86,  0.87,  0.84) & (0.94, 0.94, 0.94)\\
            A2 & (0.05, 0.04, 0.04) &  (0.19 , 0.26, 0.14) & (0.79, 0.71, 0.80) & (0.08, 0.05, 0.07) &  (1.16,  1.01,  1.06) & (0.96, 0.93, 0.92)\\
            A3 & (0.08, 0.06, 0.07) &  (0.17 , 0.15, 0.12) & (0.82, 0.82, 0.85) & (0.13, 0.12, 0.18) &  (0.93,  0.80,  0.79) & (0.94, 0.87, 0.84)\\
            A4 & (0.05, 0.04, 0.04) &  (0.23 , 0.17, 0.23) & (0.81, 0.80, 0.74) & (0.05, 0.04, 0.03) &  (1.08,  0.97,  0.98) & (0.97, 0.92, 0.96)\\
            A5 & (0.07, 0.06, 0.06) &  (0.19 , 0.15, 0.15) & (0.84, 0.83, 0.82) & (0.11, 0.12, 0.12) &  (1.02,  0.78,  0.88) & (0.95, 0.87, 0.89)\\
            \hline \hline
    \end{tabular}}
    \label{res:invivoI_tab1}
\end{table*}

\begin{figure*}[!ht]
    \centering
     \includegraphics[width=\linewidth, alt={Qualitative comparison of velocity $V_x$ for subject B1 between LR, HR and SR in peak systolic frames and correlation plots between SR and HR, covering all voxels within the left ventricle and aorta.}]{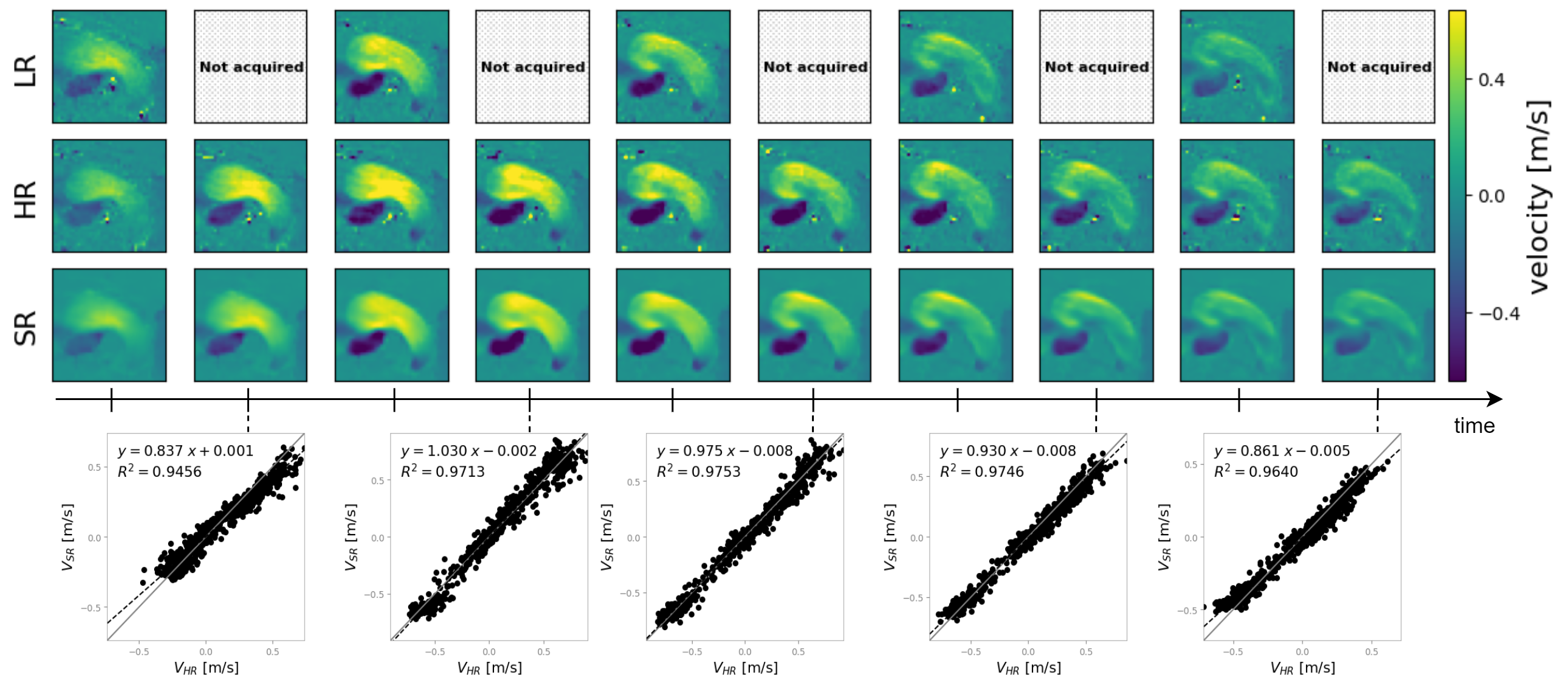}
    \caption{Qualitative comparison of velocity $V_x$ for subject B1 between LR, HR and SR in peak systolic frames and correlation plots between SR and HR, including points from the entire fluid region of the left ventricle and aorta.}
    \label{fig:res_invivoII_qualitative_timeseries}
\end{figure*}

\subsection{In-vivo evaluation}
As presented in Section \ref{sec:methods_invivo_verification}, the network performance is further evaluated on two in-vivo datasets: one seeking to recover synthetically removed temporal frames, and one comparing datasets acquired at two different temporal resolutions. 

\subsubsection{Recovery to native resolution}
\label{sec:results_invivoI_eval}
Fig. \ref{fig:results_invivoI_qual_P01} presents a sequence of cross-sectional slices from a single subject across multiple time frames during peak systole. As visually apparent, the network effectively identifies and enhances fluid regions while suppressing noise in surrounding static tissue regions.
Further, in the synthetically generated high-resolution frames, the network can reproduce features only seen in the high-resolution reference data. Voxel-wise linear regression comparison across the peak systolic frames highlights strong agreement between super-resolved and high-resolution data, with an average linear regression coefficient of $k = 0.93$ and an average coefficient of determination of $R^2 = 0.89$. 
Table \ref{res:invivoI_tab1} shows the same data across all evaluated subjects for both the entire cardiac cycle, as well as for a synthesized frame during peak systole. Here, slightly larger deviations are observed across subjects, however, recovery maintains at an average RMSE of 4.9 cm/s across the cardiac cycle, or 10 cm/s in peak systole.

\begin{figure*}[!ht]
    \centering
    \includegraphics[width=\linewidth, alt={Mean velocity through planes at both ascending (\textbf{A}) and descending aorta (\textbf{B}) comparing LR, HR and SR of subject B1, B3 and B5. Velocities are scaled by heartbeat duration.}]{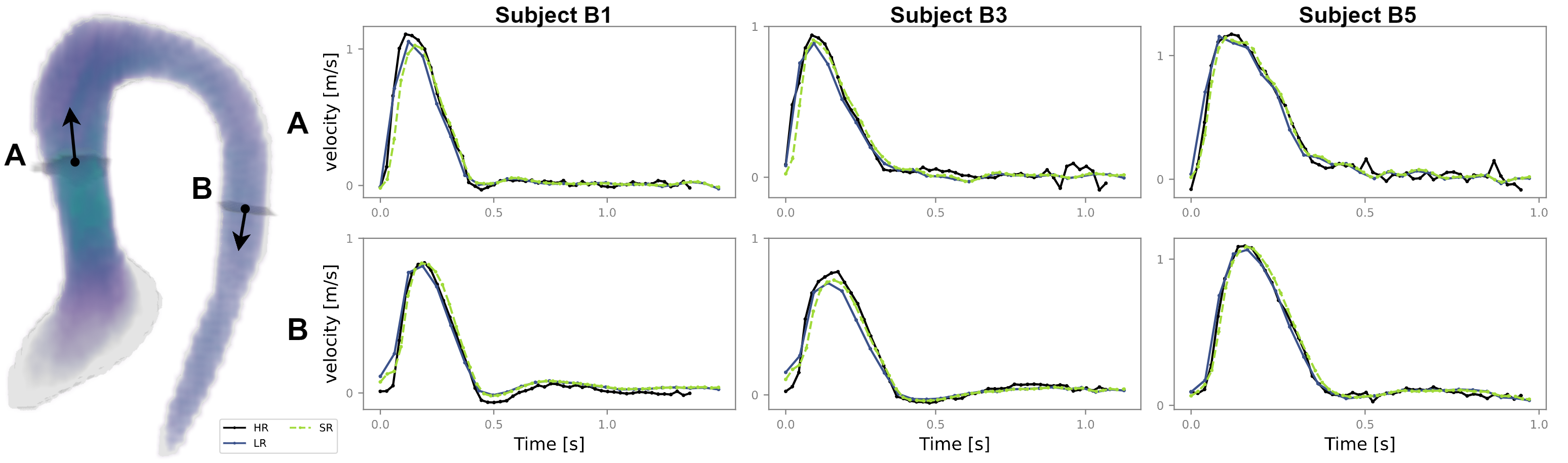}
    \caption{Mean velocity through planes at both ascending (\textbf{A}) and descending aorta (\textbf{B}) comparing LR, HR and SR of volunteer B1, B3 and B5. 
     Each plane has a thickness of around 8-10 mm to account for pixel shifts and mimic the slice thickness in 2D flow measurements of aortic flow profiles. Velocities are scaled by heartbeat duration.
    }
    \label{fig:res_invivoII_plane_flow}
\end{figure*}

\begin{table*}[t]
    \centering
    \caption{Evaluation of network performance on paired \emph{in-vivo} dataset at peak synthesized systolic time point in the aorta and in peak synthesized early diastolic frame in the left ventricle. For evaluation, the whole domain of the aorta or left ventricle is considered.}
    \resizebox{\textwidth}{!}{
        \begin{tabular}{lcccccc}
            \hline
             & \multicolumn{3}{c}{Aorta} & \multicolumn{3}{c}{Left ventricle} \\
               & RMSE  & $k$& $R^2$ &  RMSE & $k$& $R^2$ \\
   \cmidrule(lr){2-4} \cmidrule(lr){5-7} 
                                        B1 & (0.06, 0.06, 0.08) & (0.93, 0.91, 0.88) & (0.98, 0.95, 0.97) & (0.05, 0.11, 0.15) & (0.87, 0.77, 0.66) & (0.66, 0.79, 0.72)  \\
                                        B2 & (0.10, 0.08, 0.09) & (1.14, 1.11, 1.08) & (0.97, 0.94, 0.96) & (0.05, 0.06, 0.10) & (0.78, 0.74, 1.04) & (0.70, 0.82, 0.82)  \\
                                        B3 & (0.08, 0.07, 0.09) & (0.92, 0.88, 0.91) & (0.98, 0.94, 0.96) & (0.06, 0.08, 0.14) & (0.59, 0.80, 0.69) & (0.43, 0.73, 0.69)  \\
                                        B4 & (0.12, 0.10, 0.10) & (1.07, 0.95, 0.98) & (0.96, 0.88, 0.95) & (0.06, 0.08, 0.09) & (0.50, 0.76, 0.93) & (0.39, 0.76, 0.85)  \\
                                        B5 & (0.07, 0.07, 0.10) & (1.01, 0.90, 0.94) & (0.98, 0.96, 0.98) & (0.06, 0.07, 0.12) & (0.82, 0.82, 0.76) & (0.77, 0.84, 0.78)  \\ \hline \hline
        \end{tabular}
    }
    \label{tab:res_invivoII_all_volunteers}
\end{table*}

\subsubsection{Paired data acquisition}
\label{sec:results_invivoII_eval}
Fig.  \ref{fig:res_invivoII_qualitative_timeseries} depicts low, high, and super-resolved data in a thoracic cross-sectional slice during peak systole. Similar to Section \ref{sec:results_invivoI_eval}, the super-resolution network maintains the ability to identify fluid regions, suppressing signal in surrounding tissue and non-fluid regions and qualitatively enhancing flow regions. Linear regression plots highlight strong agreement between synthesized super-resolved and high-resolution reference data, with $k = 0.93$ and $R^2 = 0.97$ in peak synthesized systole. 
\noindent
Quantitative results for all subjects are summarized in Table \ref{tab:res_invivoII_all_volunteers}. With data presented for peak flow events; peak systole in the aortic domain; and peak early diastole (E-wave) in the left ventricular domain, the network achieves overall strong agreement with reference data. For the aorta in particular, an RMSE of 9 cm/s, $k = 0.90$, and $R^2=0.95$ across all subjects highlight strong agreement with only minor deviations. For left ventricular flows, measures are given at RMSE = 9 cm/s, $k = 0.72$, and $R^2 = 0.70$, indicating similar levels of absolute agreement, however, with slightly larger deviations. 
\noindent
To complement the above, local evaluations of regional aortic flows are also depicted in Fig. \ref{fig:res_invivoII_plane_flow}. As presented,  the temporal super-resolution network does not completely mitigate differences between low- and high-resolution data, although similar features are observed. 

\section{Discussion}
In this study, we implemented a residual neural network for temporal super-resolution of 4D Flow MRI. Evaluation across both \emph{in-silico} and \emph{in-vivo} datasets highlights the model's ability to accurately synthesize realistic temporal data, with performance maintained across various patient anatomies and spatiotemporal resolutions.
\subsection{In-silico performance}
\emph{In-silico} evaluation indicated excellent agreement between synthesized super-resolved and reference high-resolution velocities. Notably, performance appeared stable across the cardiac cycle, with only slightly higher errors in frames with rapidly changing flow. Synthesized velocity data showed higher errors along fluid-domain boundaries, consistent with findings from prior studies \cite{Ferdian2023CerebrovascularMRI}. These discrepancies are likely exacerbated by the anatomical motion throughout the time series. \\
Compared to deterministic approaches, super-resolution showed improvement over both linear and sinc interpolations, both with higher errors, particularly along domain boundaries, again consistent with previous results \cite{Ferdian2023CerebrovascularMRI},\cite{Ferdian20204DFlowNet:Dynamics}. Notably, the results show how interpolation methods are substantially impacted by the lack of denoising, which is automatically incorporated into our super-resolution network.
\\
In terms of comparative performance, our approach was not evaluated against other learned alternatives. Although head-to-head comparisons are of value, alternative approaches for temporal super-resolution 4D Flow MRI remain scarce. Many existing methods are either challenging to implement due to their complexity or are not publicly available, making validation cumbersome. Instead, efforts were concentrated on internal validation through tuning hyperparameters, depth and layer design, as well as modifying the original data matching loss from 4DFlowNet \cite{Ferdian20204DFlowNet:Dynamics}. 
\subsection{CFD to 4D Flow MRI - including realistic MRI features}
A central component of our work lies in the pipeline for converting simulated patient-specific CFD data to synthetic training data resembling clinical 4D Flow MRI data. While previous approaches have used k-space cropping to reduce spatial resolution \cite{Ferdian20204DFlowNet:Dynamics} \cite{Shit2022SRflow:Data}, we included sequence-specific k-space sampling patterns along with a compressed sensing reconstruction to ensure that reconstruction characteristics and temporal artifacts observed at lower resolution propagated into the training data. In our preliminary analysis, we observed improvements in \emph{in-vivo} performance when comparing a network trained on synthetic 4D Flow MRI generated by our pipeline, compared to a network trained on data generated by merely removing every second time frame and adding complex noise in k-space. 
Specifically, three distinct aspects were visually improved: (1) identification of fluid regions in 4D Flow MRI data; (2) suppression of noise in surrounding tissue regions, and (3) reconstruction of peak flow velocities with mitigation of excessive temporal smoothing. Herein, performance can be attributed both to the realistic temporal downsampling represented by the CFD-to-MRI pipeline, as well as by the inclusion of realistic magnitude data. 
\subsection{Clinical application - translating to \emph{in-vivo}}
\label{sec:discussion_invivo_results}
To highlight the potential for clinical translation, our temporal super-resolution approach was evaluated across $n=10$ clinically acquired 4D Flow MRI datasets. As shown in Sections \ref{sec:results_invivoI_eval}-\ref{sec:results_invivoII_eval}, the network was able to successfully generalize to unseen \emph{in-vivo} data increasing temporal resolution by two, i.e. below 20 ms. Notably, across both evaluation sets exhibiting different spatial resolutions, volume orientations, and temporal resolutions, the network was consistently able to correctly identify and enhance the fluid region of not just the left heart, but also other flow compartments, such as the right heart and the aorta. Herein, the evaluation on paired \emph{in-vivo} data represents a particularly important test bed, showcasing how time frames absent in the low-resolution data can be effectively synthesized by our suggested network approach. 
\\
It is worth noting that evaluated metrics are slightly lower in the \emph{in-vivo} setup as compared to the \emph{in-silico} equivalent. The high-resolution \emph{in-vivo}  data does \emph{not} represent an ideal noise-free ground truth, but is itself subject to noise and possible acquisition artifacts. This effect is exacerbated at lower flow rates, where differences in mean vs. peak performance (Table \ref{res:invivoI_tab1} left vs. right) are partially due to noise impacting the quality of the reference high-resolution data in the diastolic frames. \\
Finally, even if the clinical utility of temporal super-resolution 4D Flow MRI remains to be determined, numerous potential applications could benefit from improved temporal resolution, enhancing data quantification or physiological insight. Current consensus statements suggest a temporal resolution of $<40$ ms for general 4D Flow MRI \cite{Dyverfeldt20154DStatement}. However, the same guidelines highlight that acquisitions at 40 ms could lead to biased flow rates and flow volumes compared to reference 2D Flow acquisitions. Phantom studies have also shown how aortic peak flows, mean velocities, and stroke volumes measured by 4D Flow converge with 2D Flow reference data only at a temporal resolution approaching 20 ms. Moreover, in regions with high flow, temporal resolution has a greater impact on accuracy than spatial resolution \cite{Montalba2018VariabilityPhantom}. Studies on pulse wave velocity have also indicated that temporal resolutions under 30 ms are necessary for realistic estimates \cite{Dorniak2016RequiredTonometry}. While spatial resolution remains a primary determinant of accuracy in narrow vasculatures \cite{Ferdian2023CerebrovascularMRI}, studies on carotid anatomies have highlighted direct effects on temporal resolution on flow rates \cite{ElSayed2023OptimizationBifurcation}. Enhanced temporal resolution has the potential to provide added physiological insights, particularly in rapid flow observed across stenotic or regurgitant valves, hypertrophic outflow obstructions, or coarcted vessel sections, where rapid hemodynamic changes play a critical role.
\subsection{Temporal super-resolution networks - establishing a baseline}
By developing a convolutional residual network for \emph{temporal} super-resolution 4D Flow MRI, we have developed a first-of-a-kind network that can generalize to unseen regions and does not depend on pre-defined flow region segmentation. Whereas most previous attempts in the 4D Flow MRI domain have focused exclusively on spatial upsampling \cite{Ferdian20204DFlowNet:Dynamics}, \cite{Ferdian2023CerebrovascularMRI}, \cite{ Long2023Super-resolutionLearning}, \cite{Shit2022SRflow:Data}, networks that enable temporal upsampling have so far been based on coordinate-mappings, requiring either patient-specific retraining or \emph{a-priori} boundary conditions  \cite{Fathi2020Super-resolutionNets} \cite{Saitta2024ImplicitMRI}. Improved generalizations of coordinate-based approaches have been suggested outside of the medical domain \cite{Jagtap2020ExtendedEquations} \cite{Lee2023Locality-AwareRepresentation} but have yet to reach clinical implementation. 
\\
Temporal super-resolution has been well-studied for natural videos, with long short-term memory (LSTM) \cite{Hochreiter1997LongMemory} and vision transformers \cite{Lu2022VideoTransformer} showcased as particularly suited for time-series analysis. In comparison, our residual neural network is not specifically designed to handle time-series, and should instead be viewed as a proof-of-principle extending already existing work in the field of 4D Flow MRI into the less-explored temporal domain. Noteworthy, preliminary assessments of our setup indicate that a convolutional LSTM did \emph{not} improve network performance, although, comprehensive evaluation requires future studies. Overall, our work could be viewed as an effective baseline model and data processing pipeline, with open-source publishing enabling benchmarking and comparison of possible future developments.
\subsection{Limitations}
\label{sec:discussion_limitations}
A few limitations are worth pointing out. First, only a limited number of datasets were used for training and testing, with six different \emph{in-silico} models representing the training data set. While spatiotemporal patches greatly increase the number of training data points, including other left heart configurations, or adding data from other cardiovascular compartments could improve performance. With that in mind, the extensive \emph{in-vivo} evaluation show maintained clinical performance and represents added validation of the synthetic training data. 
Future research will focus on adding more \emph{in-vivo} test data to better verify our method.
\\
Second, although the validation on \emph{in-vivo} data is a major component of our work, the analysis has its challenges. When acquiring pairs of high-SNR, high-resolution 4D Flow MRI data, differences in scan time, subject movement, heart rate, SNR, etc. might influence the comparison. Care was taken when acquiring the paired data, and although voxel-wise comparisons provide valuable indications of general performance, inherent differences might still exist between de-noised, super-resolved data and the high-resolution comparison. Still, the accuracy in resolving peak flow frames in Section \ref{sec:results_invivoII_eval} highlights the robustness of our evaluated approach.  
\\
Third, by adapting an existing spatial super-resolution network from 3D to 2D+t, some spatial information is discarded in the input patches. Averaging of predictions across all directions is used to mitigate this effect, however, could average out information only present in one spatial direction. Modifications for simultaneous spatiotemporal upsampling through higher-order convolutions could allow for improved spatial coherency, which could be a focus of future studies.
\section{Conclusion}
This study presents a temporal super-resolution network that effectively increases temporal resolution and denoises 4D Flow MRI, with focus on left ventricular flows. Clinical utility was demonstrated by validating the network on clinically acquired datasets at varying spatiotemporal resolutions. In particular, the network's ability to generalize across various patient anatomies without further training increases clinical applicability.

\section*{Acknowledgment}
\noindent
We thank the National Academic Infrastructure for Super-computing in Sweden for computational resources at the National Supercomputer Centre at Linköping University and Ning Jin, PhD (Siemens Medical Solutions USA, Inc.) for providing the 4D Flow research sequence.
\bibliography{amain}

\begin{thebibliography}{10}

\bibitem{Richter2006CardiologyFlow}
Yoram Richter and Elazer~R. Edelman.
\newblock {Cardiology is flow}.
\newblock {\em Circulation}, 113(23):2679--2682, 6 2006.

\bibitem{Bolger2007TransitResonance}
Ann~F. Bolger, Einar Heiberg, Matts Karlsson, Lars Wigstr{\"{o}}m, Jan Engvall, Andreas Sigfridsson, Tino Ebbers, John Peder~Escobar Kvitting, Carl~Johan Carlh{\"{a}}ll, and Bengt Wranne.
\newblock {Transit of blood flow through the human left ventricle mapped by cardiovascular magnetic resonance}.
\newblock {\em Journal of Cardiovascular Magnetic Resonance}, 9(5):741--747, 9 2007.

\bibitem{Markl20124DMRI}
Michael Markl, Alex Frydrychowicz, Sebastian Kozerke, Mike Hope, and Oliver Wieben.
\newblock {4D Flow MRI}.
\newblock {\em Journal of Magnetic Resonance Imaging}, 2012.

\bibitem{Zhuang2021ThePerspectives}
Baiyan Zhuang, Arlene Sirajuddin, Shihua Zhao, and Minjie Lu.
\newblock {The role of 4D flow MRI for clinical applications in cardiovascular disease: current status and future perspectives}.
\newblock {\em Quantitative Imaging in Medicine and Surgery}, 11(9):4193, 9 2021.

\bibitem{Ha20194DValves}
Hojin Ha, John Peder~Escobar Kvitting, Petter Dyverfeldt, and Tino Ebbers.
\newblock {4D Flow MRI quantification of blood flow patterns, turbulence and pressure drop in normal and stenotic prosthetic heart valves}.
\newblock {\em Magnetic Resonance Imaging}, 55:118--127, 1 2019.

\bibitem{Vasanawala2015CongenitalMRI}
Shreyas~S. Vasanawala, Kate Hanneman, Marcus~T. Alley, and Albert Hsiao.
\newblock {Congenital heart disease assessment with 4D flow MRI}.
\newblock {\em Journal of Magnetic Resonance Imaging}, 42(4):870--886, 10 2015.

\bibitem{Ferdian2023CerebrovascularMRI}
Edward Ferdian, David Marlevi, J~Schollenberger, M~Aristova, E~R Edelman, S~Schnell, C~A Figueroa, D~A Nordsletten, A~A Young, and D~Al Marlevi.
\newblock {Cerebrovascular super-resolution 4D Flow MRI-using deep learning to non-invasively quantify velocity, flow, and relative pressure Cerebrovascular super-resolution 4D Flow MRI}.
\newblock {\em Medical Image Analysis}, 88:102831, 2023.

\bibitem{Stankovic2014ReproducibilityResolution}
Zoran Stankovic, Bernd Jung, Jeremy Collins, Maximilian~F. Russe, James Carr, Wulf Euringer, Lena Stehlin, Zoltan Csatari, Peter~C. Strohm, Mathias Langer, and Michael Markl.
\newblock {Reproducibility study of four-dimensional flow MRI of arterial and portal venous liver hemodynamics: Influence of spatio-temporal resolution}.
\newblock {\em Magnetic Resonance in Medicine}, 72(2):477--484, 8 2014.

\bibitem{Carlsson2011QuantificationT}
Marcus Carlsson, Johannes T{\"{o}}ger, Mikael Kanski, Karin~Markenroth Bloch, Freddy St{\aa}hlberg, Einar Heiberg, and Håkan Arheden.
\newblock {Quantification and visualization of cardiovascular 4D velocity mapping accelerated with parallel imaging or k-t BLAST: Head to head comparison and validation at 1.5 T and 3 T}.
\newblock {\em Journal of Cardiovascular Magnetic Resonance}, 13(1):1--7, 10 2011.

\bibitem{Hanneman2014MagneticImaging}
Kate Hanneman, Milani Sivagnanam, Elsie~T. Nguyen, Rachel Wald, Andreas Greiser, Andrew~M. Crean, Sebastian Ley, and Bernd~J. Wintersperger.
\newblock {Magnetic resonance assessment of pulmonary (QP) to systemic (QS) flows using 4D phase-contrast imaging: pilot study comparison with standard through-plane 2D phase-contrast imaging}.
\newblock {\em Academic radiology}, 21(8):1002--1008, 2014.

\bibitem{Fyrdahl2020Sector-wiseDysfunction}
Alexander Fyrdahl, Joao~G. Ramos, Maria~J. Eriksson, Kenneth Caidahl, Martin Ugander, and Andreas Sigfridsson.
\newblock {Sector-wise golden-angle phase contrast with high temporal resolution for evaluation of left ventricular diastolic dysfunction}.
\newblock {\em Magnetic Resonance in Medicine}, 83(4):1310--1321, 4 2020.

\bibitem{Leuprecht2003BloodStudy}
Armin Leuprecht, Sebastian Kozerke, Peter Boesiger, and Karl Perktold.
\newblock {Blood flow in the human ascending aorta: A combined MRI and CFD study}.
\newblock {\em Journal of Engineering Mathematics}, 47(3-4):387--404, 12 2003.

\bibitem{Bonini2022HemodynamicRegurgitation}
M.~Bonini, M.~Hirschvogel, Y.~Ahmed, H.~Xu, A.~Young, P.C. Tang, and D.~Nordsletten.
\newblock {Hemodynamic Modeling for Mitral Regurgitation}.
\newblock {\em The Journal of Heart and Lung Transplantation}, 41(4):S218, 4 2022.

\bibitem{Dyverfeldt20154DStatement}
Petter Dyverfeldt, Malenka Bissell, Alex~J. Barker, Ann~F. Bolger, Carl~Johan Carlh{\"{a}}ll, Tino Ebbers, Christopher~J. Francios, Alex Frydrychowicz, Julia Geiger, Daniel Giese, Michael~D. Hope, Philip~J. Kilner, Sebastian Kozerke, Saul Myerson, Stefan Neubauer, Oliver Wieben, and Michael Markl.
\newblock {4D flow cardiovascular magnetic resonance consensus statement}.
\newblock {\em Journal of Cardiovascular Magnetic Resonance}, 17(1), 8 2015.

\bibitem{Dong2014ImageNetworks}
Chao Dong, Chen~Change Loy, Kaiming He, and Xiaoou Tang.
\newblock {Image Super-Resolution Using Deep Convolutional Networks}.
\newblock {\em IEEE Transactions on Pattern Analysis and Machine Intelligence}, 38(2):295--307, 12 2014.

\bibitem{Saitta2024ImplicitMRI}
Simone Saitta, Marcello Carioni, Subhadip Mukherjee, Carola-Bibiane Sch{\"{o}}nlieb, and Alberto Redaelli.
\newblock {Implicit neural representations for unsupervised super-resolution and denoising of 4D flow MRI}.
\newblock {\em Computer Methods and Programs in Biomedicine}, (246):108057, 4 2024.

\bibitem{Fathi2020Super-resolutionNets}
Mojtaba~F. Fathi, Isaac Perez-Raya, Ahmadreza Baghaie, Philipp Berg, Gabor Janiga, Amirhossein Arzani, and Roshan~M. D'Souza.
\newblock {Super-resolution and denoising of 4D-Flow MRI using physics-Informed deep neural nets}.
\newblock {\em Computer Methods and Programs in Biomedicine}, 197, 12 2020.

\bibitem{Ferdian20204DFlowNet:Dynamics}
Edward Ferdian, Avan Suinesiaputra, David~J. Dubowitz, Debbie Zhao, Alan Wang, Brett Cowan, and Alistair~A. Young.
\newblock {4DFlowNet: Super-Resolution 4D Flow MRI Using Deep Learning and Computational Fluid Dynamics}.
\newblock {\em Frontiers in Physics}, 8:138, 5 2020.

\bibitem{Shit2022SRflow:Data}
Suprosanna Shit, Judith Zimmermann, Ivan Ezhov, Johannes~C. Paetzold, Augusto~F. Sanches, Carolin Pirkl, and Bjoern~H. Menze.
\newblock {SRflow: Deep learning based super-resolution of 4D-flow MRI data}.
\newblock {\em Frontiers in Artificial Intelligence}, 5:171, 8 2022.

\bibitem{Long2023Super-resolutionLearning}
Derek Long, Cameron McMurdo, Edward Ferdian, Charlène~A. Mauger, David Marlevi, Martyn~P. Nash, and Alistair~A. Young.
\newblock {Super-resolution 4D flow MRI to quantify aortic regurgitation using computational fluid dynamics and deep learning}.
\newblock {\em International Journal of Cardiovascular Imaging}, pages 1--14, 2 2023.

\bibitem{Ericsson2024GeneralizedSystem}
Leon Ericsson, Adam Hjalmarsson, Muhammad~Usman Akbar, Edward Ferdian, Mia Bonini, Brandon Hardy, Jonas Schollenberger, Maria Aristova, Patrick Winter, Nicholas Burris, Alexander Fyrdahl, Andreas Sigfridsson, Susanne Schnell, C.~Alberto Figueroa, David Nordsletten, Alistair~A. Young, and David Marlevi.
\newblock {Generalized super-resolution 4D Flow MRI - using ensemble learning to extend across the cardiovascular system}.
\newblock {\em IEEE journal of biomedical and health informatics}, PP, 2024.

\bibitem{Kingma2015Adam:Optimization}
Diederik~P Kingma and Jimmy Lei~Ba.
\newblock {Adam: A Method for Stochastic Optimization}.
\newblock In {\em International Conference on Learning Representations}, 2015.

\bibitem{Forman2014High-resolutionCompensation}
Christoph Forman, Davide Piccini, Robert Grimm, Jana Hutter, Joachim Hornegger, and Michael~O. Zenge.
\newblock {High-resolution 3D whole-heart coronary MRA: a study on the combination of data acquisition in multiple breath-holds and 1D residual respiratory motion compensation}.
\newblock {\em Magnetic Resonance Materials in Physics, Biology and Medicine}, 27(5), 2014.

\bibitem{Lustig2007SparseImaging}
Michael Lustig, David Donoho, and John~M. Pauly.
\newblock {Sparse MRI: The application of compressed sensing for rapid MR imaging}.
\newblock {\em Magnetic Resonance in Medicine}, 58(6):1182--1195, 12 2007.

\bibitem{Beck2009AProblems}
Amir Beck and Marc Teboulle.
\newblock {A fast iterative shrinkage-thresholding algorithm for linear inverse problems}.
\newblock {\em SIAM Journal on Imaging Sciences}, 2(1), 2009.

\bibitem{Uecker2015BerkeleyToolbox}
Martin Uecker, F~Ong, JI~Tamir, D~Bahri, P~Virtue, JY~Cheng, T~Zhang, and M.~Lustig.
\newblock {Berkeley Advanced Reconstruction Toolbox}.
\newblock In {\em ISMRM}, 2015.

\bibitem{Montalba2018VariabilityPhantom}
Cristian Montalba, Jesus Urbina, Julio Sotelo, Marcelo~E. Andia, Cristian Tejos, Pablo Irarrazaval, Daniel~E. Hurtado, Israel Valverde, and Sergio Uribe.
\newblock {Variability of 4D flow parameters when subjected to changes in MRI acquisition parameters using a realistic thoracic aortic phantom}.
\newblock {\em Magnetic Resonance in Medicine}, 79(4):1882--1892, 4 2018.

\bibitem{Dorniak2016RequiredTonometry}
Karolina Dorniak, Einar Heiberg, Marcin Hellmann, Dorota Rawicz-Zegrzda, Maria Wesierska, Rafal Galaska, Agnieszka Sabisz, Edyta Szurowska, Maria Dudziak, and Erik Hedstr{\"{o}}m.
\newblock {Required temporal resolution for accurate thoracic aortic pulse wave velocity measurements by phase-contrast magnetic resonance imaging and comparison with clinical standard applanation tonometry}.
\newblock {\em BMC Cardiovascular Disorders}, 16(1):1--9, 5 2016.

\bibitem{ElSayed2023OptimizationBifurcation}
Retta El~Sayed, Alireza Sharifi, Charlie~C. Park, Diogo~C. Haussen, Jason~W. Allen, and John~N. Oshinski.
\newblock {Optimization of 4D Flow MRI Spatial and Temporal Resolution for Examining Complex Hemodynamics in the Carotid Artery Bifurcation}.
\newblock {\em Cardiovascular Engineering and Technology}, 14(3):476--488, 6 2023.

\bibitem{Jagtap2020ExtendedEquations}
Ameya~D. Jagtap and George~Em Karniadakis.
\newblock {Extended physics-informed neural networks (XPINNs): A generalized space-time domain decomposition based deep learning framework for nonlinear partial differential equations}.
\newblock {\em Communications in Computational Physics}, 28(5):2002--2041, 11 2020.

\bibitem{Lee2023Locality-AwareRepresentation}
Doyup Lee, Chiheon Kim, Minsu Cho, and Wook-Shin Han.
\newblock {Locality-Aware Generalizable Implicit Neural Representation}.
\newblock {\em Advances in Neural Information Processing Systems}, 36:48363--48381, 12 2023.

\bibitem{Hochreiter1997LongMemory}
Sepp Hochreiter and Jürgen Schmidhuber.
\newblock {Long Short-Term Memory}.
\newblock {\em Neural Computation}, 9(8):1735--1780, 11 1997.

\bibitem{Lu2022VideoTransformer}
Liying Lu, Ruizheng Wu, Huaijia Lin, Jiangbo Lu, and Jiaya Jia.
\newblock {Video Frame Interpolation With Transformer}.
\newblock In {\em 2022 IEEE/CVF Conference on Computer Vision and Pattern Recognition}, pages 3522--3532, 2022.

\end{thebibliography}

\end{document}